\documentclass[a4paper,conference]{IEEEtran}
\usepackage{cite}

\ifCLASSINFOpdf
  \usepackage[pdftex]{graphicx}
\else
\fi
\usepackage{amsmath}

\usepackage{stfloats}
\usepackage{hyperref}
\hypersetup{colorlinks=true,linkcolor=blue,urlcolor=cyan}

\usepackage{multirow}

\begin{document}
%
\title{IPN Hand: A Video Dataset and Benchmark for Real-Time Continuous Hand Gesture Recognition}




%
\author{\IEEEauthorblockN{
Gibran Benitez-Garcia \IEEEauthorrefmark{1},
Jesus Olivares-Mercado\IEEEauthorrefmark{2},
Gabriel Sanchez-Perez \IEEEauthorrefmark{2}
and Keiji Yanai\IEEEauthorrefmark{1}}
\IEEEauthorblockA{\IEEEauthorrefmark{1}Department of Informatics, The University of Electro-Communications, Tokyo, Japan\\Email: gibran@ieee.org, yanai@cs.uec.ac.jp}
\IEEEauthorblockA{\IEEEauthorrefmark{2}Instituto Politecnico Nacional, ESIME Culhuacan, Mexico City, Mexico\\Email: jolivares@ipn.mx, gasanchezp@ipn.mx}}


\maketitle

\begin{abstract}
Continuous hand gesture recognition (HGR) is an essential part of human-computer interaction with a wide range of applications in the automotive sector, consumer electronics, home automation, and others. In recent years, accurate and efficient deep learning models have been proposed for HGR. However, in the research community, the current publicly available datasets lack real-world elements needed to build responsive and efficient HGR systems. In this paper, we introduce a new benchmark dataset named IPN Hand with sufficient size, variety, and real-world elements able to train and evaluate deep neural networks. This dataset contains more than 4,000 gesture samples and 800,000 RGB frames from 50 distinct subjects. We design 13 different static and dynamic gestures focused on interaction with touchless screens. We especially consider the scenario when continuous gestures are performed without transition states, and when subjects perform natural movements with their hands as non-gesture actions. Gestures were collected from about 30 diverse scenes, with real-world variation in background and illumination. With our dataset, the performance of three 3D-CNN models is evaluated on the tasks of isolated and continuous real-time HGR. Furthermore, we analyze the possibility of increasing the recognition accuracy by adding multiple modalities derived from RGB frames, i.e., optical flow and semantic segmentation, while keeping the real-time performance of the 3D-CNN model. Our empirical study also provides a comparison with the publicly available nvGesture (NVIDIA) dataset. The experimental results show that the state-of-the-art ResNext-101 model decreases about 30\% accuracy when using our real-world dataset, demonstrating that the IPN Hand dataset can be used as a benchmark, and may help the community to step forward in the continuous HGR. Our dataset and pre-trained models used in the evaluation are publicly available at \href{github.com/GibranBenitez/IPN-hand}{github.com/GibranBenitez/IPN-hand}.


\end{abstract}


%
\IEEEpeerreviewmaketitle

\section{Introduction}
\begin{figure}[t]
  \begin{center}
  \begin{tabular}[c]{ccc}
    \multicolumn{3}{c}{\includegraphics[width=0.98\linewidth]{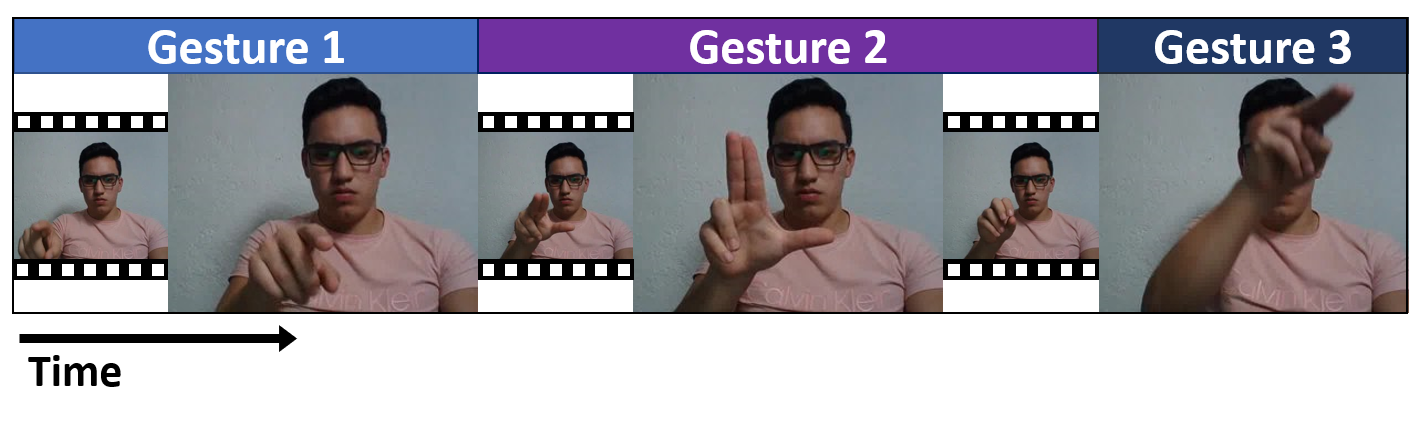}} \vspace{-2em}\\
    \multicolumn{3}{c}{(a)} \vspace{1mm} \\
    \multicolumn{3}{c}{\includegraphics[width=0.98\linewidth]{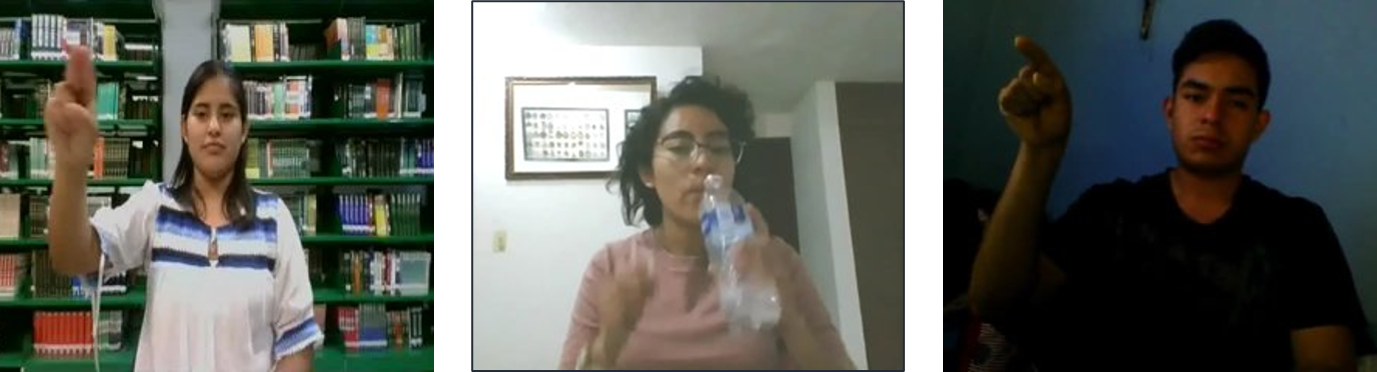}} \\
    \multicolumn{3}{l}{\hspace{11mm} (b) \hspace{23mm} (c) \hspace{23mm} (d)} \\\\
  \end{tabular}
  \vspace{-2.5em}
  \end{center}
\caption{
Some examples that show the challenges of our dataset.
(a) Continuous gestures without transition states.
(b) Clutter backgrounds.
(c) Natural interaction with objects.
(d) Weak illumination conditions.}
\label{fig:problems}
\end{figure}

Gestures are a natural form of human communication~\cite{kendon1980}.
Hand gesture recognition (HGR) is an essential part of human-computer interaction.
Systems using vision-based interaction and control are more common nowadays~\cite{leo2017,berg2017,rautar2015}.
Compared to traditional inputs such as mouse and keyboard, vision-based interfaces can be more practical and natural based on the intuitiveness of designed gestures.
HGR has a wide range of applications in the automotive sector, consumer electronics, public
transit, gaming, home automation, and others~\cite{picke2007,parada2014hand,rautar2015}.
For these applications, HGR systems must be designed to work online and deal with continuous gestures that users may input. 
There are mainly four issues that systems must deal with for continuous HGR applications:
(i) continuous gestures without transition states, 
(ii) natural behaviors of users' hands similar to target gestures, 
(iii) intra-class variability of gestures' duration, 
and (iv) lag between performing a gesture and its classification.
It is worth mentioning that (i) is particularly important for some real-world applications such as interacting with touchless screens. 
For example, someone can open a picture and zoom in, by using two continuous gestures (double-click + zoom-in) without a transition state that forces the hand returning to a neutral position before the second gesture. 
This example is shown in Figure~\ref{fig:problems}(a).

Thanks to the advances on deep neural networks, in recent years, accurate and efficient deep learning models have been proposed to overcome the challenges on continuous HGR~\cite{molchanov2016online,liu2017continuous,zhu2018continuous,narayana2019continuous,kopuklu2019real}.
However, these methods were evaluated using datasets that do not cover the main real-world issues.
Currently, it is not easy to find a benchmark dataset able to evaluate the four main issues of continuous HGR.
Most of the hand gesture dataset used for continuous HGR, like ChaLearn ConGD~\cite{chalearn2016}, nvGesture~\cite{molchanov2016online} and EgoGesture~\cite{2018egogesture}, do not include continuous gestures without transition states (i), nor natural hand movements as non-gesture actions (ii). 
To the best of our knowledge, there a no publicly available hand gesture datasets that cover these two issues of continuous HGR. 
Note that, some works have designed specific datasets for controlling automotive interfaces~\cite{benitez2019mva,benitez2020sens}, that partially include these issues. 
However, the datasets are not publicly available.

In this paper, we introduce a new dataset called IPN Hand for the task of continuous hand gesture recognition.
The dataset contains more than four thousand RGB gesture samples and 800 thousand frames from 50 distinct subjects.
We design 13 classes of static and dynamic gestures for interaction with touchless screens. 
Our dataset has the most realistic scenario for continuous HGR than other hand gesture datasets. 
IPN Hand includes the largest number of continuous gestures per video and the largest speed of intra-class variation for different subjects when they were performing the same gesture.
Besides, our dataset is more complex as it was collected from about thirty representative scenes with considerable variation, including clutter backgrounds, strong and weak illumination conditions, static and dynamic background environments. 
We specially design two scenarios that reflects the real-world issues of continuous HGR: when continuous gestures are performed without transition states, and when subjects perform natural movements with their hands as non-gesture actions.
Some examples of the main challenges of our dataset are shown in Figure~\ref{fig:problems}.

Given our dataset, we evaluate three state-of-the-art (SOTA) methods based on 3D-CNN models for the tasks of isolated and continuous real-time HGR.
Furthermore, we analyze the possibility of increasing the recognition accuracy by adding multiple modalities derived from RGB frames. 
Specifically, we evaluate the data level fusion of RGB with semantic segmentation of target hands as an alternative of the RGB+Optical Flow or RGB+Depth modalities for real-time HGR.
Our empirical study also provides a comparison with the publicly available nvGesture (NVIDIA) dataset.
The experimental results on the IPN Hand dataset demonstrate that it can be used as a benchmark for the data-hungry 3D-CNN methods, which may help the community to step forward in the continuous HGR.

\section{Related Works}
\subsection{Datasets for continuous HGR}
Existing continuous HGR datasets differ by factors such as scale, the number of classes, sensors used, and the domain of gestures. 
However, all of them must include non-gesture frames to emulate the online behavior of real applications.
These frames are important to define the realism of the designed dataset. Therefore, we analyze the commonly used datasets based on this element.

\begin{table*}[t]
\caption{Comparison of the public continuous gesture datasets}
\label{tab:datasets}
\centering
\begin{tabular}{lcccccccc}
\hline
Dataset              & Instances & Videos & Instance/video & Classes & Subjects & Scenes & View & Modalities \\
\hline
ChaLearn ConGD, 2016~\cite{chalearn2016} & 47,933    & 22,535 & 2.1            & 249     & 21       & 15     & 3rd  & RGB-D      \\
nvGesture, 2016~\cite{molchanov2016online}     & 1,532     & 1,532  & 1.0            & 25      & 20       & 1      & 3rd  & RGB-D      \\
EgoGesture, 2018~\cite{2018egogesture}    & 24,161    & 2,081  & 11.6           & 83      & 50       & 6      & 1st  & RGB-D      \\
IPN Hand (ours)      & 4,218     & 200    & 21.1           & 13      & 50       & 28     & 3rd  & RGB       \\
\hline
\end{tabular}
\end{table*}

The ChaLearn LAP ConGD dataset~\cite{chalearn2016} is derived from 9 different gesture domains, from Italian sign language, activities to pantomime.
It contains 249 classes and more than 40 thousand instances from 21 subjects, which makes ChaLearn ConGD the largest dataset for continuous HGR. 
This dataset contains videos with one to five continuous gestures with transition states.
The nvGesture dataset~\cite{molchanov2016online} is designed to control in-car automotive devices, and it includes 25 gesture types from 20 subjects, consisting of 1 532 instances. 
Note that this dataset is commonly used for online HGR evaluation, even though it only contains videos with isolated gestures.
The EgoGesture dataset~\cite{2018egogesture} is a benchmark dataset for egocentric (first-person) view, which consists of 83 classes with more than 20 thousand instances from 50 subjects.
This dataset includes at max 12 instances per video.
To the best of our knowledge, our proposed IPN Hand dataset contains the largest number of instances per video (21), and different scenes (28), making it suitable for evaluating continuous HGR systems. 
Detailed comparison between our IPN Hand and related gesture datasets
can be found in Table~\ref{tab:datasets}.

\begin{figure}[t]
  \begin{center}
  \begin{tabular}[c]{cccc}
    \multicolumn{4}{c}{\includegraphics[width=0.98\linewidth]{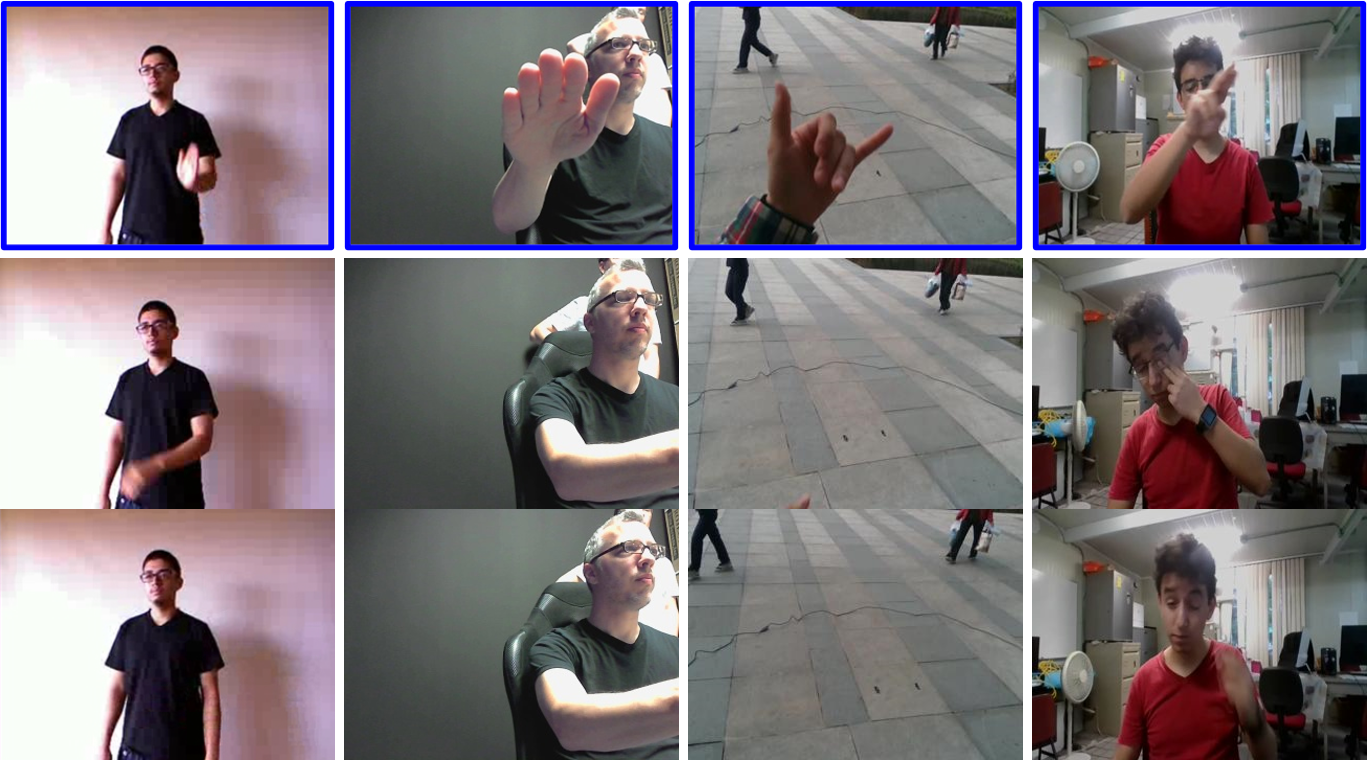}} \\
    \multicolumn{4}{l}{\small \hspace{4mm} ConGD \hspace{9mm} nvGesture \hspace{5mm} EgoGesture \hspace{6mm} IPN Hand} \\
  \end{tabular}
  \vspace{-1em}
  \end{center}
\caption{
Comparison of gesture vs. non-gesture frames from the public continuous gesture datasets.
Note that only the first row (blue) shows examples of gesture frames.}
\label{fig:no_gest}
\end{figure}

Figure~\ref{fig:no_gest} shows the comparison of the non-gesture frames from the public continuous gesture datasets.
Most of the non-gesture frames from the ChaLearn ConGD dataset consist of transition states between gestures, usually showing the subject in a neutral position as in the first column of the figure. On the other hand, the nvGesture shows drivers with their hands at the steering wheel, while EgoGesture shows only backgrounds with the hands out of view as non-gesture frames.
It is clear that our dataset presents more challenge on distinguishing gesture vs. non-gesture frames, since we include subjects performing natural movements with their hands.

\subsection{Continuous HGR}
Continuous HGR can be divided into two stages: gesture spotting and gesture classification. 
Gesture spotting aims to detect temporal video segments which contain gesture instances, while the classification stage aims to classify the gesture of each spotted segment.
Multi-stream architectures have been widely employed for both tasks. 
Simonyan et al.~\cite{simonyan2014two} were the pioneers of fusing features from two modalities, using one stream with RGB images and the other with flow fields for isolated HGR.
On 
This multi-modality approach is prevalent as well for continuous HGR, as shown in the 2017 ChaLearn Look At People (LAP) continuous gesture detection (ConGD) challenge~\cite{wan2017results},
where all entries used multi-stream architectures of at least RGB and depth. 
For instance, the winners~\cite{liu2017continuous} introduced a two-stream 3D-CNN combining hand-location features of RGB and depth modalities by explicitly overlaying a black mask on the input frames.
They firstly spotted the gestures based on a dataset-specific observation: subjects raise their hands at the beginning of gestures and put them down again at the end. 
One year after ConGD challenge, Zhu et al.~\cite{zhu2018continuous} overcame the results of the winners by proposing a temporal dilated 3D-CNN architecture to binary classify gesture/non-gesture frames, and 3D-CNN+LSTM for gesture classification.
The current SOTA method of ConGD uses a 12-channels architecture with extra RNN layers to simultaneously spot and classify continuous gestures~\cite{narayana2019continuous}.
Recently, Kopuklu et al.~\cite{kopuklu2019real} proposed a hierarchical structure of 3D-CNN architectures to detect and classify continuous hand gestures.
Their spotting method consists of detecting gesture frames using a shallow 3D-CNN model on eight consecutive frames, while a SOTA model is used for classification only if a gesture is detected. 

To evaluate our dataset for real-world applications of continuous HGR, detection and classification must work online or even with a negative lag between performing a gesture and its classification feedback.
Therefore, we follow the hierarchical structure of 3D-CNNs~\cite{kopuklu2019real}, since it can detect a gesture when a confidence measure reaches a threshold level before the gesture ends (early-detection).
Furthermore, we evaluate the multi-modality accuracy derived from RGB frames, by keeping in mind the real-time performance. 
We employ a data level fusion to avoid the significant increase in the computational cost of the 3D-CNN models.
Besides, we propose to use semantic segmentation results as an alternative to the absent depth modality and the computational expensive optical flow.

\begin{figure}[t]
    \begin{center}
  \begin{tabular}[c]{ccc}
    \multicolumn{3}{c}{\includegraphics[width=0.98\linewidth]{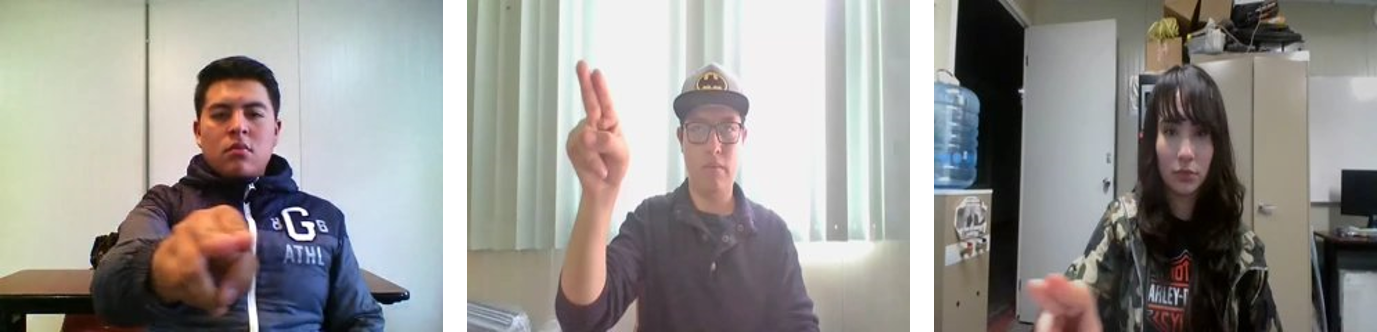}} \\
    \multicolumn{3}{l}{\small \hspace{6mm} (a) 1.Plain \hspace{11mm} (b) 2.WindowA \hspace{8mm} (c) 3.OfficeA } \vspace{1mm}\\
    \multicolumn{3}{c}{\includegraphics[width=0.98\linewidth]{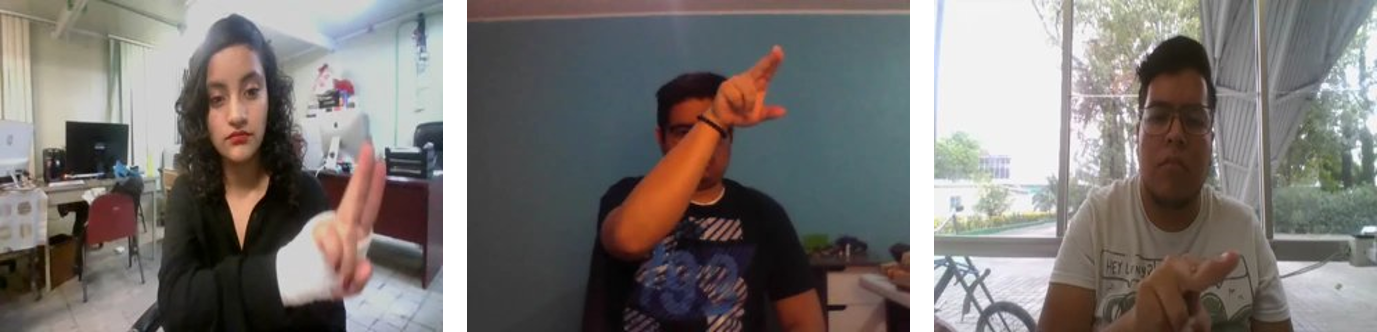}} \\
    \multicolumn{3}{l}{\small \hspace{4mm} (d) 4.OfficeB \hspace{7.5mm} (e) 5.DarkRoom \hspace{6mm} (f) 6.WindowB }\\
  \end{tabular}
  \vspace{-1em}
  \end{center}
\caption{
Some examples of the scenes with more videos in the dataset.
Note that all subjects are developing the same gesture, "pointing with two fingers".}
\label{fig:scenes}
\end{figure}

\section{The IPN Hand Dataset}

\subsection{Data collection}
To collect the dataset, the RGB videos were recorded in the resolution of $640 \times 480$ with the frame rate of 30 fps. 
The participants were asked to record the gestures using their own PC or laptop by keeping the defined resolution and frame rate. 
Thus, the distance between the camera and each subject varies, since we instructed all participants to be located in a comfortable position to manipulate the screen of their PC with hand gestures.
In this way, the videos were generated with 28 different scenes in total.
For some participants that do not have access to a camera that can cover the recording specifications, we prepared three scenes (scene 2-4, shown in Figure~\ref{fig:scenes}(b)-(d)), including clutter and variable illumination backgrounds.
When collecting data, we first teach the subjects how to perform each gesture and tell them the gesture names (short descriptions). 
Then we generate four gesture name lists with random order for recording four videos per subject.
Thus, the subject was told the gesture name and performed the gesture accordingly.
They were asked to continuously perform 21 gestures with 3 random breaks as a single session which was recorded as a video.

\subsection{Dataset characteristics}

\subsubsection{Gesture Classes}
\begin{figure*}[t]
\centering
\includegraphics[width=1\textwidth]{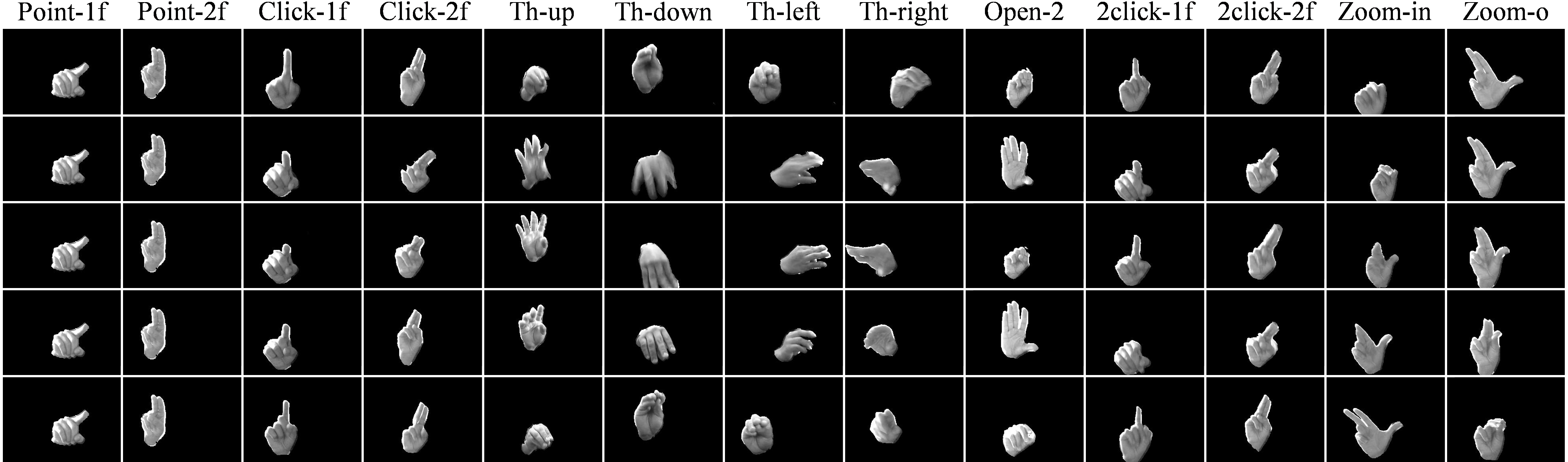}
\caption{Examples of the 13 gesture classes included in our dataset.
For visualization purposes, semantic segmentation masks were blended to the RGB images.
}
\label{fig:data}
\end{figure*}

We design the gestures in our dataset focused on interaction with touchless screens.
We include gestures able to control the location of the pointer on the screen (pointer), and to manipulate interfaces (actions).
We take some gestures used in common smartphones, as they are natural and easy to remember by the users. 
Thus, we designed 13 gestures (shown in Figure~\ref{fig:data}) defined to control the punter and actions.

For the punter location, we defined two static gestures of pointing to the screen with one and two fingers, respectively.
Note that these gestures are static due to the gesture itself does not need temporal information to be detected. However, the application of this gesture includes the hand movement to control the pointer.
For the action gestures, we defined 11 gestures, including click with one and two fingers, throw to four positions (up, down, left \& right), double click with one and two fingers, zoom-in, and zoom-out.
Table~\ref{tab:stats} provides the name and descriptions of each gesture in our dataset.
Note that we also include segments were natural hand movements are performed as non-gestures states.

\begin{table}[tbh]
\caption{Statics per gesture of our IPN Hand dataset}
\label{tab:stats}
\centering
\begin{tabular}{clcccc}
\hline
id & \multicolumn{1}{c}{Gesture}    &  Name     & Instances   & \begin{tabular}[c]{@{}c@{}}Mean\\duration\\($std$)\end{tabular} & \begin{tabular}[c]{@{}c@{}}Duration\\ $\overline{nstd}_k$\end{tabular} \\
\hline 
\vspace{-2mm}
\\
0  & No gesture                                                              & No-gest   & 1431      & 147 (133)                                                    & 0.904                                                      \vspace{1.5mm}\\
1  & \begin{tabular}[c]{@{}l@{}}Pointing with\\ one finger\end{tabular}      & Point-1f  & 1010      & 219 (67)                                                     & 0.308                                                      \vspace{1.5mm}\\
2  & \begin{tabular}[c]{@{}l@{}}Pointing with\\ two fingers\end{tabular}     & Point-2f  & 1007      & 224 (69)                                                     & 0.309                                                      \vspace{1.5mm}\\
3  & \begin{tabular}[c]{@{}l@{}}Click with\\ one finger\end{tabular}         & Click-1f  & 200       & 56 (29)                                                      & 0.517                                                      \vspace{1.5mm}\\
4  & \begin{tabular}[c]{@{}l@{}}Click with\\ two fingers\end{tabular}        & Click-2f  & 200       & 60 (43)                                                      & 0.718                                                      \vspace{1.5mm}\\
5  & Throw up                                                                & Th-up     & 200       & 62 (25)                                                      & 0.400                                                      \vspace{1.5mm}\\
6  & Throw down                                                              & Th-down   & 201       & 65 (28)                                                      & 0.424                                                      \vspace{1.5mm}\\
7  & Throw left                                                              & Th-left   & 200       & 66 (27)                                                      & 0.400                                                      \vspace{1.5mm}\\
8  & Throw right                                                             & Th-right  & 200       & 64 (28)                                                      & 0.439                                                      \vspace{1.5mm}\\
9  & Open twice                                                              & Open-2    & 200       & 76 (31)                                                      & 0.410                                                      \vspace{1.5mm}\\
10 & \begin{tabular}[c]{@{}l@{}}Double click\\ with one finger\end{tabular}  & 2click-1f & 200       & 68 (28)                                                      & 0.412                                                      \vspace{1.5mm}\\
11 & \begin{tabular}[c]{@{}l@{}}Double click\\ with two fingers\end{tabular} & 2click-2f & 200       & 70 (30)                                                      & 0.435                                                      \vspace{1.5mm}\\
12 & Zoom in                                                                 & Zoom-in   & 200       & 65 (29)                                                      & 0.440                                                      \vspace{1.5mm}\\
13 & Zoom out                                                                & Zoom-o    & 200       & 64 (28)                                                      & 0.432                                                     \vspace{1mm}\\
\hline
\end{tabular}
\end{table}

\subsubsection{Subjects}
A small number of subjects could make the intra-class variation very limited. 
Hence, we asked 50 subjects for our data collection. 
In the 50 subjects, there are 16 females and 34 males. 
The average age of the subjects is 21.2, where the minimum age is 18, and the maximum age is 25. 
All of the subjects are currently students at Instituto Politecnico from Mexico.

\subsubsection{Extreme illumination and clutter backgrounds}
To evaluate the robustness to the illumination change of the baseline methods, we have data
collected under extreme conditions such as facing a window with strong sunlight 
(Figure~\ref{fig:scenes}(b)), or in a room with almost null artificial light (Figure~\ref{fig:problems}(d)).
Some scenes with static background placed with student-life stuff (Figure~\ref{fig:problems}(b)), and dynamic background with walking people appearing in the camera view (Figure~\ref{fig:scenes}(d)-(f)) were also included.

\begin{figure}[t]
\centering
\includegraphics[width=\linewidth]{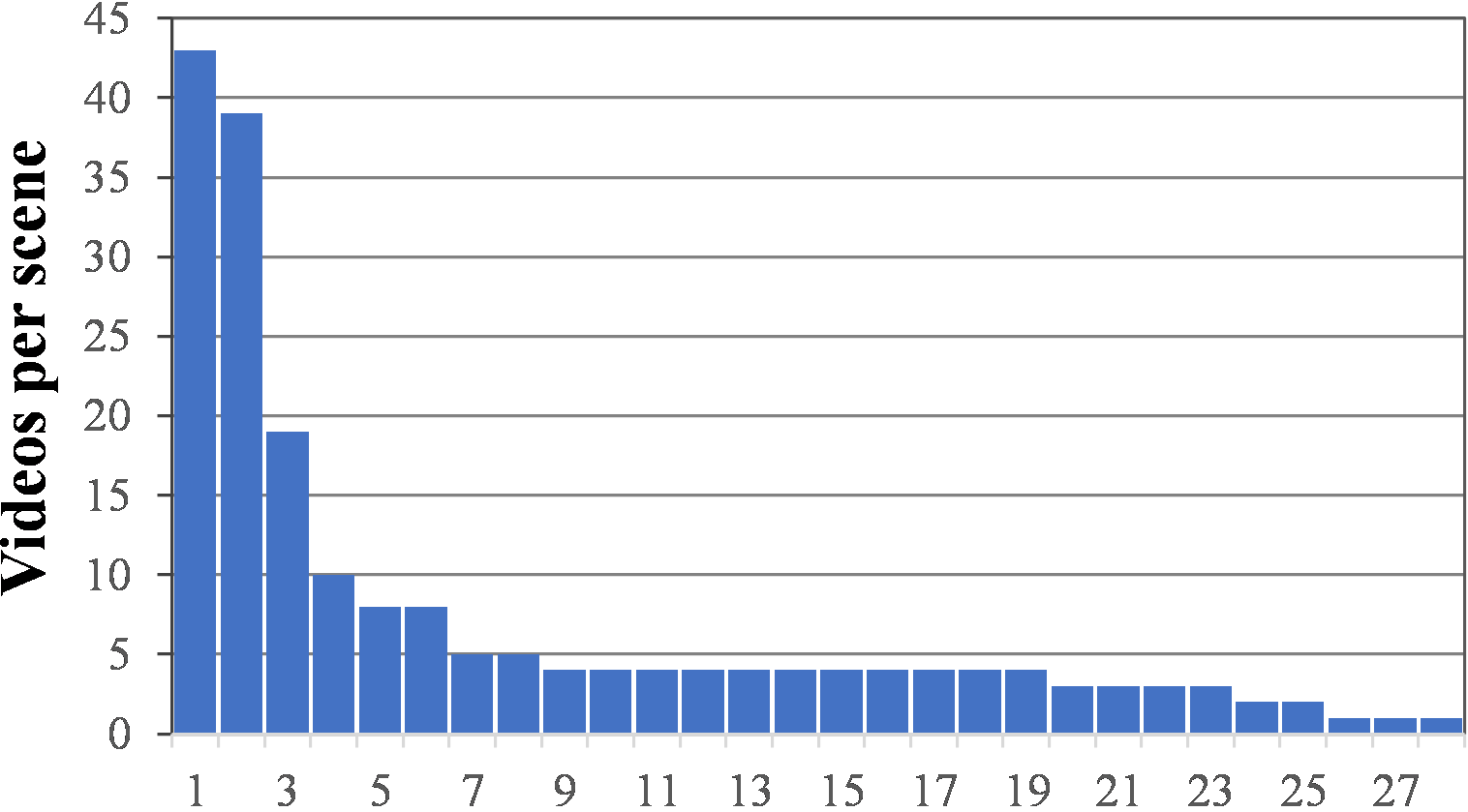}
\caption{Distribution of the 28 different scenes included in the dataset.}
\label{fig:freq}
\end{figure}

\subsection{Dataset statics}

Fifty distinct subjects participated in performing 13 classes of gestures in 28 diverse scenes. 
Totally 4,218 gesture instances and 800,491 frames were collected in RGB. 
Figure~\ref{fig:freq} illustrates the sample distribution on each video among the 28 scenes. 
In the figure, the horizontal axis and the vertical axis indicate the ID of the scenes and the number of the videos, respectively. 
In addition, Figure~\ref{fig:scenes} shows examples of the six scenes with more videos on our dataset. 

When data collection, 21 gestures are considered as a session and recorded in a single video. 
Thus, we get 200 RGB videos in total. 
Note that the order of the gestures performed is semi-randomly generated, trying to include one dynamic after one static gesture, which resembles the realistic interaction with a touchless screen.
The start and end frame index of each gesture instance in the video were manually labeled, which provides enough information to train and evaluate continuous HGR approaches in a fully supervised manner.
In the dataset, the minimum length of a gesture is 9 frames. The maximum length of a gesture is 650
frames. 

\begin{table*}[t]
\caption{Statics of the public continuous gesture datasets}
\label{tab:staDat}
\centering
\begin{tabular}{lcccccc}
\hline
Dataset              & Frames    & Mean video duration & Mean gesture duration & Duration std & Duration $\overline{nstd}_k$ & \% of train \\
\hline
ChaLearn ConGD, 2016~\cite{chalearn2016} & 1,714,629 & 76.1                & 41                    & 18.5         & 0.37           & 0.635       \\
nvGesture, 2016~\cite{molchanov2016online}      & 122,560   & 80.0                & 71                    & 32.3         & 0.3            & 0.685       \\
EgoGesture, 2018~\cite{2018egogesture}       & 2,953,224 & 1,419.1             & 38                    & 13.9         & 0.33           & 0.595       \\
IPN Hand (ours)      & 800,491   & 4,002.5             & 140                   & 93.9         & 0.43           & 0.739    \\
\hline
\end{tabular}
\end{table*}

In Table~\ref{tab:staDat}, we show the data statistics of our dataset compared with the continuous HGR datasets that are currently publicly available.
The dataset statistics include the number of total frames, the mean video durations, the mean of the gesture sample durations, the standard deviation of gesture durations, the percentage of the training data. The mean and standard deviation of the gesture sample durations is calculated over the samples from all gesture classes in the dataset.
Following~\cite{2018egogesture}, we use the normalized standard deviation for gesture duration in each gesture class to describe the speed variation of different subjects when performing the same gesture. The normalized standard deviation of durations in a gesture class $k$ is calculated as follows:
\begin{eqnarray}
nstd_k = \frac{1}{l^{\bar{k}}} 
\sqrt[]{\frac{\sum_i^{N}(l_i^k-l^{\bar{k}})^2}{N}}
\label{eq:nstdk}
\end{eqnarray}
where in gesture class $k$, $l_i^k$ represents the duration of the $i$th sample, $l^{\bar{k}}$ is the average duration of samples, and $N$ is the number of samples.
We get the average $\overline{nstd}_k$ over all gesture classes, for the whole dataset.  

From Table~\ref{tab:staDat}, we can see that our IPN Hand datset has the largest duration $\overline{nstd}_k$ (0.43), significantly larger than EgoGesture, which is the second largest. 
This demonstrates that our dataset has a large speed variation for different subjects when performing the same gesture.
In resume, our proposed dataset covers all the real-world issues for continuous HGR, as described in the introduction:
(i) continuous gestures without transition states, 
(ii) natural behaviors of users' hands similar to target gestures, 
(iii) intra-class variability of gestures' duration ($\overline{nstd}_k$).

\section{Benchmark Evaluation}
We evaluate three SOTA 3D-CNN models as baselines for the tasks of isolated and continuous HGR, with our new IPN Hand dataset. 
For continuous HGR, we adopt a two-model hierarchical architecture to detect and classify the continuous input stream of frames.

\begin{figure}[t]
\centering
\includegraphics[width=\linewidth]{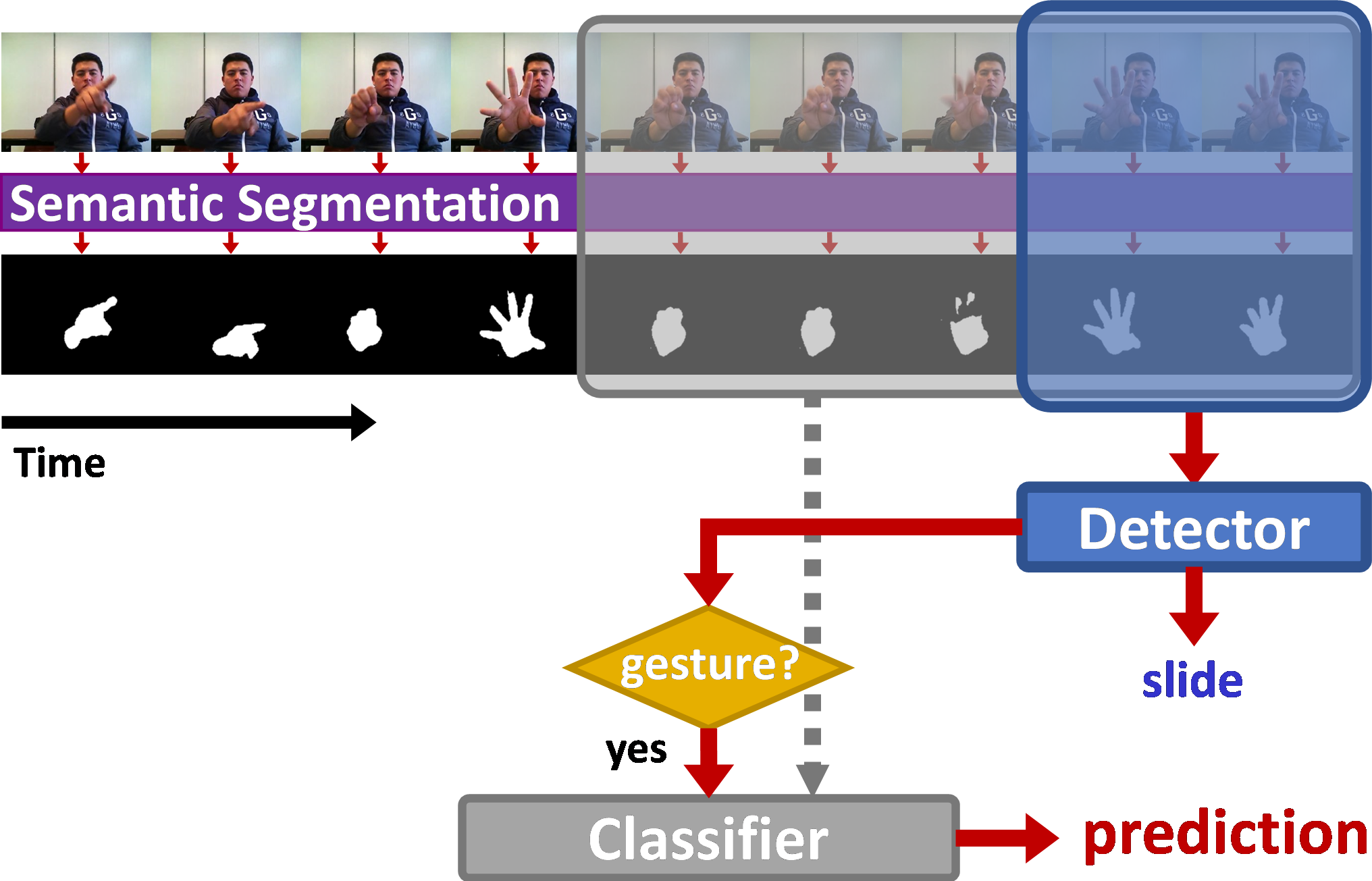}
\caption{The flowchart of the continuous HGR approach based on two hierarchical 3D-CNNs models.}
\label{fig:method}
\end{figure}

\subsection{Hierarchical 3D-CNNs for continuous HGR}
As mentioned before, we use the framework proposed by Kopuklu et al.~\cite{kopuklu2019real} to detect and classify continuous hand gestures.
The flowchart of the two-model hierarchical architecture is shown in Figure~\ref{fig:method}.
Sliding windows with a constant stride run through incoming video frames where the detector queue is placed at the very beginning of the classifier queue. 
If the detector recognizes a gesture, then the classifier is activated. 
The detector’s output is post-processed for more robust performance, and the final decision is made using a single-time activation block where only one activation occurs per performed gesture. Note that the sliding window of the detector is the only who is working every single stride. In contrast, the classification window is activated based on detectors output, so the inference time is not always the sum of both models.

Since we also evaluate multi-modal 3D-CNNs models, we first calculate the semantic segmentation masks of each streaming frame. 
For this task, we use the efficient semantic segmentation approach of HarDNet (Harmonic Dense Net)~\cite{HarDNet} which achieves SOTA results with a network inspired by DenseNet~\cite{DenseNet}. 
Its core component, the \textit{HarDBlock} (Harmonic Dense Block), is specifically designed to address the problem of the GPU memory traffic.
Therefore, we can achieve real-time performance on each frame before applying 3D-CNNs models.

\subsection{Experimental setup}
We randomly split the data by subject into training (74\%), and testing (26\%) sets, resulting in 148 training, and 52 testing videos. 
The numbers of gesture instances in training, and testing splits are 3 117, and 1 101, respectively. 
Thus, 37 and 13 subjects were designed for training and testing, respectively.

\subsubsection{Implementation details}
We compare the 3D-CNN models of C3D~\cite{c3d2015}, and the 3D versions~\cite{hara2018can} of Resnet-50~\cite{he2016deep} and ResNeXt-101~\cite{xie2017resnext}, as deep models for classification. 
On the other hand, for the shallow 3D-CNN detector we use the ResLight (Resnet-10) as proposed in~\cite{kopuklu2019real}. 
All 3D-CNN models were trained using a fully supervised strategy using the manually segmented gestures.

For the multi-modality tests, we train a HarDNet~\cite{HarDNet} model, and use the SPyNet~\cite{ranjan2017SPyNet} approach for calculation of semantic segmentation and optical flow, respectively. 
We trained the HarDNet model with a synthetic hand pose estimation dataset~\cite{zimmermann2017learning} which contains more than 40 thousand images with hands fully annotated at pixel-level\footnote{\href{https://lmb.informatik.uni-freiburg.de/projects/hand3d/}{https://lmb.informatik.uni-freiburg.de/projects/hand3d/}}.
For the SPyNet, we used the open-source implementation and pre-trained model of~\cite{pytorch-spynet} to calculate real-time densely optical flow from each input frame.

All 3D-CNN models were pre-trained on the Jester dataset~\cite{2019jester}, while the HarDNet on the ImageNet dataset~\cite{ImageNet}.
Contrary, since ResLight is a compact model, we trained it from scratch using all non-gesture and gesture instances of the IPN Hand training set.
The inference time (FPS) was measured on an Intel Core i7-9700K desktop with a single NVIDIA GTX 1080ti GPU.
More implementation details, as well as training and evaluation codes can be found in our open-source repository\footnote{\href{https://github.com/GibranBenitez/IPN-hand}{https://github.com/GibranBenitez/IPN-hand}}

\begin{table*}[t]
\caption{Results of isolated HGR task using our IPN Hand dataset}
\label{tab:isol}
\centering
\begin{tabular}{lcccccc}
\hline
Model       & Input sequence & Modality & Results        & Parameters       & Model Size      & Inference time    \\
\hline
C3D         & 32-frames      & RGB      & 77.75          & 50.75 M          & 387 MB          & 76.2 ms         \\
\hline
ResNeXt-101 & 32-frames      & RGB      & 83.59          & 47.51 M          & 363 MB          & 27.7 ms          \\
ResNeXt-101 & 32-frames      & RGB-Flow & \textbf{86.32} & 47.56 M          & 363 MB          & 28.9 ms          \\
ResNeXt-101 & 32-frames      & RGB-Seg  & 84.77          & 47.56 M          & 363 MB          & 28.9 ms          \\
ResNet-50   & 32-frames      & RGB      & 73.1           & \textbf{46.25 M} & \textbf{353 MB} & \textbf{17.8 ms} \\
ResNet-50   & 32-frames      & RGB-Flow & 74.65          & 46.27 M          & \textbf{353 MB} & 18.2 ms          \\
ResNet-50   & 32-frames      & RGB-Seg  & 75.11          & 46.27 M          & \textbf{353 MB} & 18.2 ms         \\
\hline
\end{tabular}
\end{table*}

\subsection{Isolated HGR task}
We evaluate this task with the conventional classification metric. We segment the video sequences into isolated gesture samples based on the beginning and ending frames manually annotated. The learning task is to predict class labels for each gesture sample. We use classification accuracy, which is the percent of correctly labeled examples, and the confusion matrix of the predictions, as evaluation metrics for this learning task.

\subsubsection{Experimental results using the IPN Hand dataset}

Table~\ref{tab:isol} presents the results of evaluated models with different modalities using our IPN Hand dataset. As expected, the best results were obtained by the ResNeXt-101 model, which can barely achieve real-time performance with the different modalities. However, it is clearly faster and more accurate than the robust C3D model.
On the other hand, ResNet-50 is the most efficient model, but present the lowest accuracy results among the evaluated approaches.

\begin{table}[t]
\caption{Comparison of the extra processes for multi-modality models.}
\label{tab:extra}
\centering
\begin{tabular}{lcccc}
\hline
Process      & Model   & Params.          & \begin{tabular}[c]{@{}c@{}}Model\\ size\end{tabular} & \begin{tabular}[c]{@{}c@{}}Inference\\ time\end{tabular} \\
\hline
Segmentation & HarDNet & 4.114 M          & 15.8 MB                                              & \textbf{8.1 ms}                                          \\
Optical Flow & SPyNet  & \textbf{1.440 M} & \textbf{5.50 MB}                                     & 21.9 ms      \\
\hline                                           
\end{tabular}
\end{table}

It is worth noting that our RGB-seg achieves competitive results compared to the RGB-flow, which is significant since the optical flow process is more computationally expensive.
To evaluate the efficiency of the multi-modal approaches, Table~\ref{tab:extra} shows the inference time and the model size of extra-processes related to semantic segmentation and optical flow estimation. From this table, we can see that the semantic segmentation is more than two times faster than the optical flow, making the RGB-seg alternative feasible for real-time applications.

\begin{figure}[t]
\centering
\includegraphics[width=\linewidth]{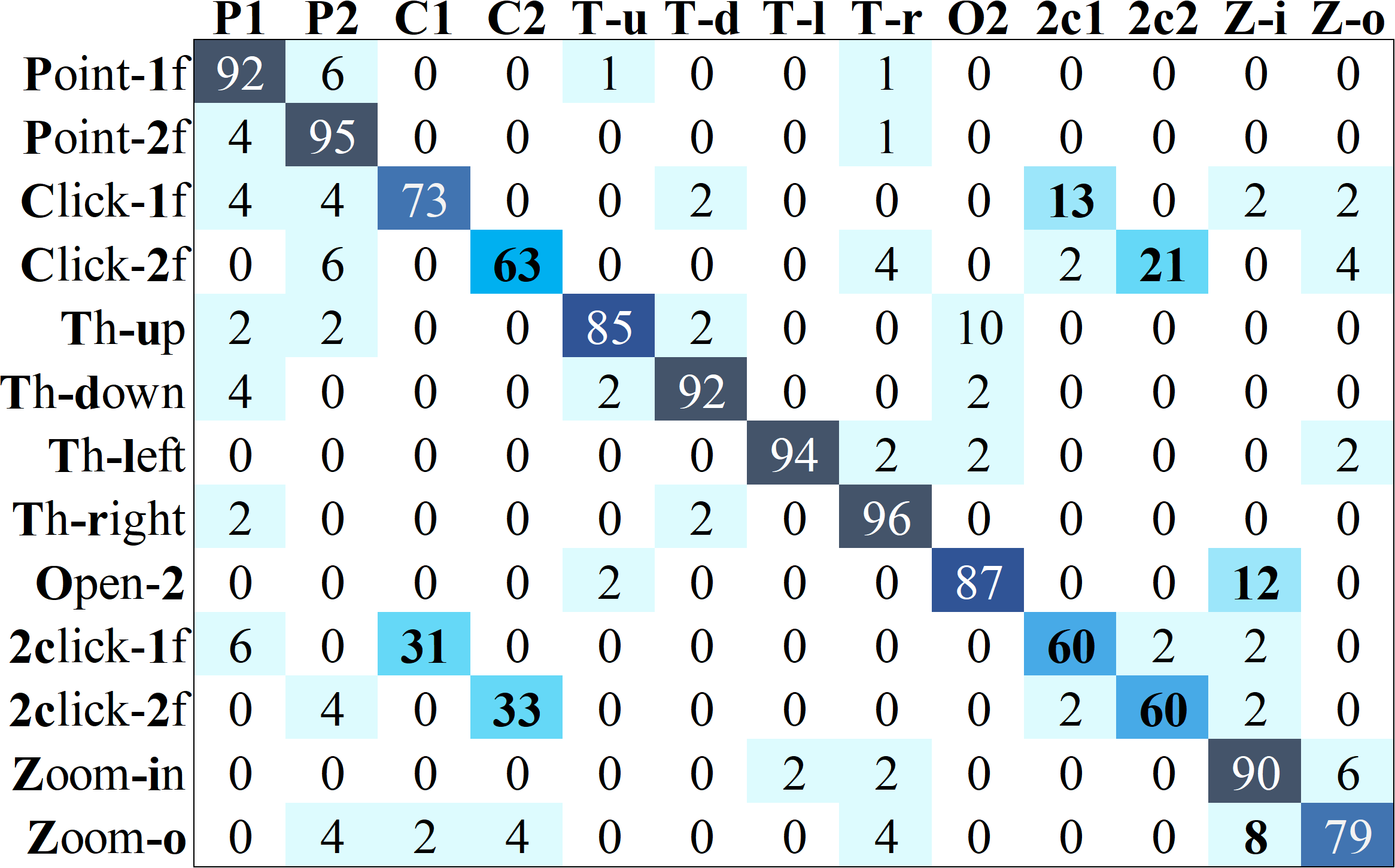}
\caption{Confusion matrix of the best result obtained from ResNext-101 model with RGB-flow.}
\label{fig:CMiso}
\end{figure}

In addition, in the Figure~\ref{fig:CMiso}, we show the confusion matrix of the best result obtained for isolated HGR, ResNext-101 with RGB-flow. As expected, the problems are related to the gestures that are closer such as clicks with double clicks.

\subsubsection{Experimental results using the nvGesture dataset}

As mentioned before, we also evaluate the performance of our RGB-seg alternative with a common online HGR dataset. Table~\ref{tab:nvSeg} presents the results of the evaluated models with different modalities using the nvGesture dataset. From these results, we can notice that the trend of the multi-modal results from IPN Hand is still valid (RGB-flow $>$ RGB-seg $>$ RGB).

\begin{table}[t]
\caption{Results of isolated HGR task using the nvGesture dataset}
\label{tab:nvSeg}
\centering
\begin{tabular}{lccc}
\hline
Model                & Input sequence     & Modality          & Results        \\
\hline
ResNeXt-101          & 32-frames          & RGB               & 79.46          \\
\textbf{ResNeXt-101} & \textbf{32-frames} & \textbf{RGB-Flow} & \textbf{82.36} \\
ResNeXt-101          & 32-frames          & RGB-Seg           & 82.15  \\
\hline
\end{tabular}
\end{table}

In addition, Table~\ref{tab:nvSota} shows the result of our real-time RGB-seg alternative compared to SOTA methods of nvGesture dataset. From this table, we validate the use of semantic segmentation as an important source of valuable features for HGR. The results of ResNeXt-101 with RGB-seg, are competitive to SOTA methods that even employ different modalities, such as depth. These findings are significant due to the semantic segmentation process only takes 8 ms with an input image of $360 \times 240$ pixels.

\begin{table}[t]
\caption{Comparision with SOTA methods of nvGesture dataset.}
\label{tab:nvSota}
\centering
\begin{tabular}{lcc}
\hline
Model        & Modality          & Results       \\
\hline
C3D~\cite{c3d2015}         & RGB               & 73.8          \\
R3DCNN~\cite{molchanov2016online}       & RGB               & 74.1          \\
MTUT~\cite{abavisani2019mtut}          & RGB*              & 81.3          \\
R3DCNN~\cite{molchanov2016online}       & RGB+Flow          & 79.3          \\
\textbf{MFF~\cite{kopu2018motion}} & \textbf{RGB+Flow} & \textbf{84.7} \\
R3DCNN~\cite{molchanov2016online}       & Depth+Flow        & 82.4          \\
ResNeXt-101  & RGB+Seg           & 82.2 \\
\hline        
\end{tabular}
\end{table}

\subsection{Continuous HGR task}
We use the Levenshtein accuracy~\cite{kopuklu2019real} as evaluation metric for the continuous HGR taks.
This metric employs the Levenshtein distance to measure the distance between sequences by counting the number of item-level changes (insertion, deletion, or substitutions) to transform one sequence into the other.
In the case of continuous hand gestures, the difference between the sequence of predicted and ground truth gestures is measured. 
For example, if a ground truth sequence is $[1,2,3,4,5,6,7,8,9]$ and predicted gestures of a video is $[1,2,7,4,5,6,6,7,8,9]$, the Levenshtein distance is 2. 
Defined by the deletion of one of the ”6” which is detected two times, and the substitution of ”7” with ”3”. 
Thus, the Levenshtein accuracy is obtained by averaging this distance over the number of true target classes. 
In our example, the accuracy is  $1-(2/9)\times100 = 77.78\%$.

We obtain the average Levenshtein accuracy over the 52 testing videos to asses the continuous HGR performance. Besides, we also evaluate the detection accuracy of the 3D-CNN detectors with different multi-modalities.
Note that detectors are trained and evaluated using isolated gesture vs. non-gesture samples. So that we use a binary classification accuracy, and the confusion matrix of the predictions, as evaluation metrics for these models.

\begin{table*}[t]
\caption{Results of the detector model using our IPN Hand dataset}
\label{tab:dets}
\centering
\begin{tabular}{lcccccc}
\hline
Model       & Input sequence & Modality & Results        & Parameters       & Model Size      & Inference time    \\
\hline
ResLight-10 & 8-frames & RGB      & 75.4           & \textbf{0.895 M} & \textbf{6.83 MB} & \textbf{2.4 ms} \\
ResLight-10 & 8-frames & RGB-Flow & \textbf{82.43} & 0.908 M          & 6.94 MB          & 2.9 ms          \\
ResLight-10 & 8-frames & RGB-Seg  & 80.06          & 0.908 M          & 6.94 MB          & 2.9 ms        \\
\hline
\vspace{-5mm}
\end{tabular}
\end{table*}

\subsubsection{Experimental results}
\begin{figure}[t]
  \begin{center}
  \begin{tabular}[c]{ccc}
    \includegraphics[width=0.32\linewidth]{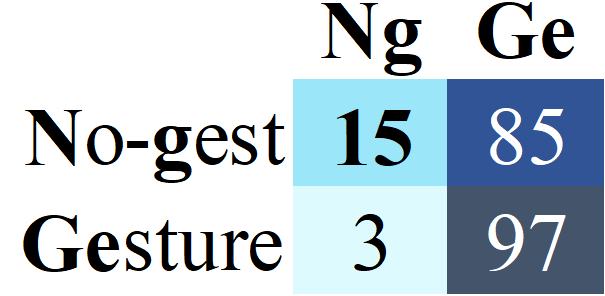} & 
    \hspace{-3mm} \includegraphics[width=0.32\linewidth]{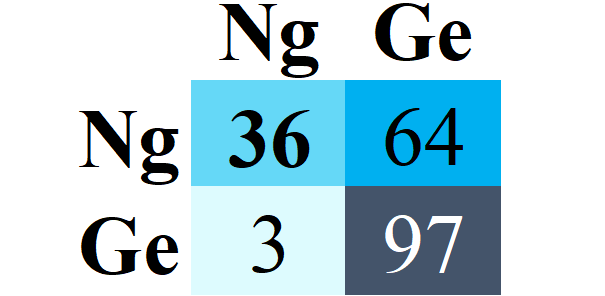} &
    \hspace{-6mm} \includegraphics[width=0.32\linewidth]{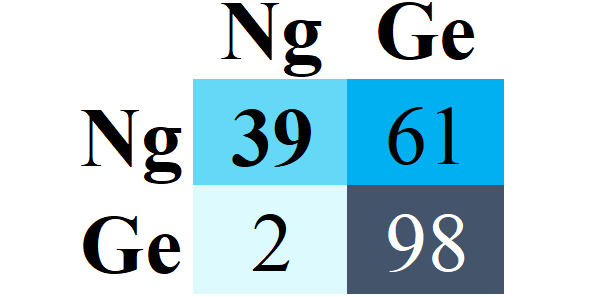} \\
    (a) RGB &
    \hspace{-2mm}(b) RGB-seg &
    \hspace{-4mm} (c) RGB-flow \\
  \end{tabular}
  \end{center}
\caption{
Confusion matrices of gesture detection on the IPN hand dataset using different modalities for the Resnet-Light model. 
}
\label{fig:CMsDet}
\end{figure}

First, we evaluate the detector model with the different modalities in Table~\ref{tab:dets}. As expected, the same trend is maintained, but since ResLight-10 is a much more compact model, the inference time is not affected. 
It is also important to analyze the misclassification of the detector, therefore Figure~\ref{fig:CMsDet} shows the confusion matrices from each modality. It is clear that the benefits from multi-modal approaches are reflected in the detection of non-gesture frames, since the RGB approach misrecognized 85\% of these frames.

\begin{figure}[t]
  \begin{center}
  \begin{tabular}[c]{c}
    \includegraphics[width=0.99\linewidth]{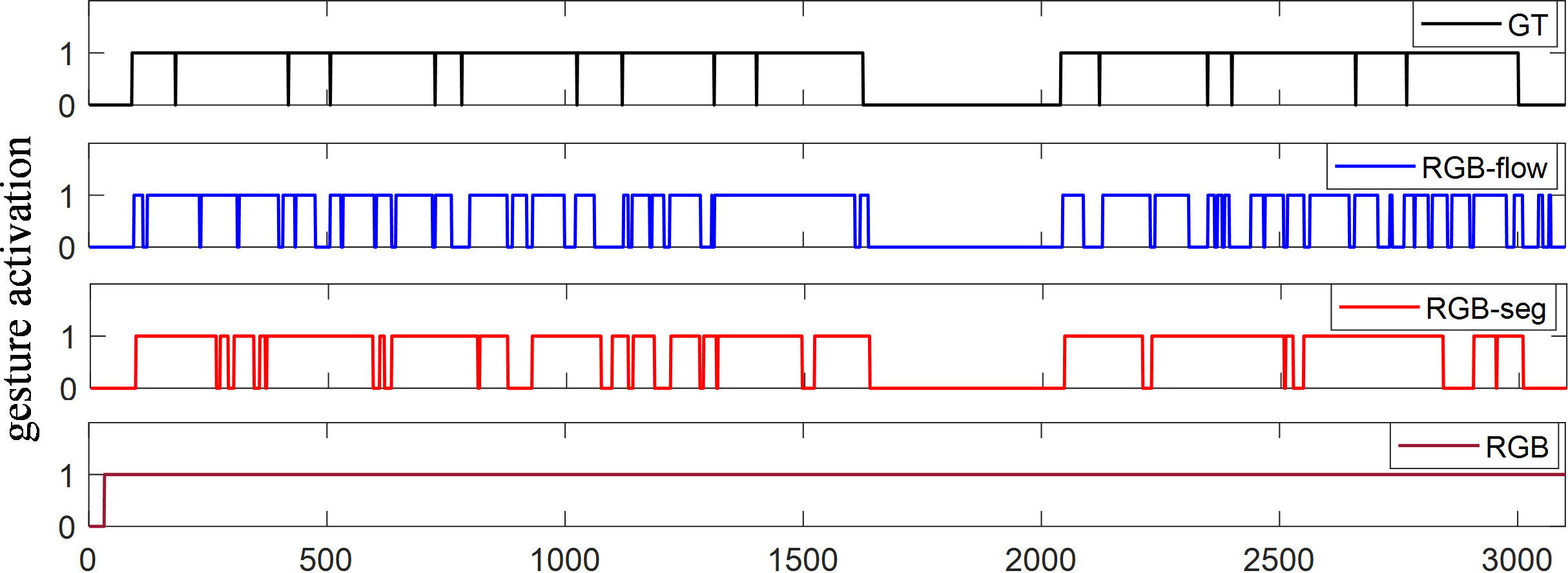}\\
    \small (a) Detection results \vspace{2mm} \\
    \includegraphics[width=0.99\linewidth]{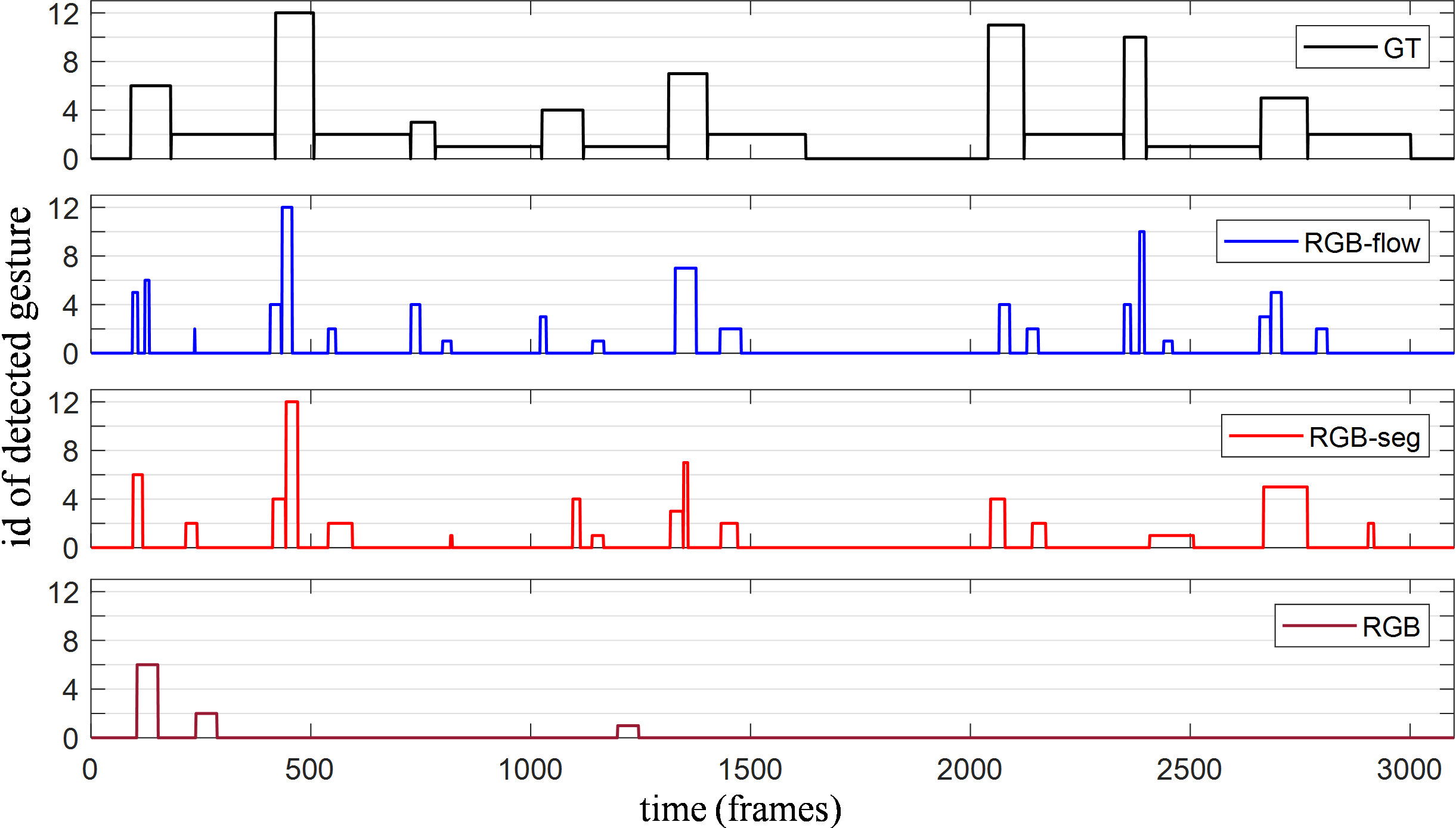} \\
    \small (b) Classification results \\
  \end{tabular}
  \vspace{-1em}
  \end{center}
\caption{Qualitative results of the proposed networks. The most significant improvements occur on pixels belonging to large objects.}
\label{fig:visual}
\end{figure}

We evaluate the complete process of the hierarchical two-model approach using Resnet-50 and ResNeXt-101 as classifiers, and ResLight-10 as a detector. In addition, we also analyze the multi-modal alternatives covering real-time performance. 
Table~\ref{tab:cont} shows the Levenshtein accuracy, as well as the model size and inference time of each evaluated model and modality. The results of this test make more evident the advantages of using the RGB-seg alternative for real-time continuous HGR. For instance, the inference time results of ResNeXt-101 RGB are comparable with those of ResNet-50 RGB-seg. However, the Levenshtein accuracy of the latter is significantly better (around 10\% of improvement).
Finally, we show the temporal visualization of continuous predictions from one testing video in Figure~\ref{fig:visual}.

\begin{table*}[t]
\caption{Results of the hierarchical two-model approach for continuous HGR using our IPN Hand dataset}
\label{tab:cont}
\centering
\begin{tabular}{lcccccccc}
\hline
            &          &                              & \multicolumn{3}{c}{Model size}                                  & \multicolumn{3}{c}{Inference time}                    \\
Model       & Modality & \multicolumn{1}{c|}{Results} & detector         & classifier      & \multicolumn{1}{c|}{total} & detector        & classifier       & total            \\
\hline
ResNeXt-101 & RGB      & 25.34                        & \textbf{6.83 MB} & \textbf{363 MB} & \textbf{370 MB}            & \textbf{2.9 ms} & \textbf{27.7 ms} & \textbf{30.1 ms} \\
ResNeXt-101 & RGB-Flow & \textbf{42.47}               & 12.4 MB          & 363 MB          & 375 MB                     & 11.1 ms         & 28.9 ms          & 53.7 ms          \\
ResNeXt-101 & RGB-seg  & 39.01                        & 22.7 MB          & 363 MB          & 386 MB                     & 24.8 ms         & 28.9 ms          & 39.9 ms          \\
Resnet-50   & RGB      & 19.78                        & \textbf{6.83 MB} & \textbf{353 MB} & \textbf{360 MB}            & \textbf{2.9 ms} & \textbf{17.8 ms} & \textbf{20.4 ms} \\
Resnet-50   & RGB-Flow & \textbf{39.47}               & 12.4 MB          & 353 MB          & 365 MB                     & 11.1 ms         & 18.2 ms          & 43.1 ms          \\
Resnet-50   & RGB-seg  & 33.27                        & 22.7 MB          & 353 MB          & 376 MB                     & 24.8 ms         & 18.2 ms          & 29.2 ms        \\
\hline
\vspace{-5mm}
\end{tabular}
\end{table*}

\section{Conclusion}
In this paper, we introduced a new benchmark dataset for continuous HGR that includes real-world issues, such as continuous gestures without transition states, natural behaviors of users' hands, and large intra-class variability of gestures' duration. 
Besides, we evaluate the data level fusion of RGB with semantic segmentation as an alternative of the RGB+Optical Flow or RGB+Depth modalities for real-time HGR.
From the experimental results, we conclude that RGB-seg is a suitable multi-modal alternative for real-time continuous hand gesture recognition. Furthermore, we believe that the proposed dataset can be used as a benchmark and help the community to move steps forward on the continuous HGR.

\section*{Acknowledgment}
This work was supported by JSPS KAKENHI Grant Number 15H05915, 17H01745, 17H06100 and 19H04929.



%

\bibliographystyle{IEEEtran}
\bibliography{my_bibliography, hgr2020}

\end{document}